\documentclass[10pt,twocolumn,letterpaper]{article}

\usepackage{ijcb}
\usepackage{times}
\usepackage{epsfig}
\usepackage{graphicx}
\usepackage{amsmath}
\usepackage{amssymb}
\usepackage{pdflscape}
\usepackage{multirow}
\usepackage{graphicx}
\usepackage[accsupp]{axessibility}

\ijcbfinalcopy 

\ifijcbfinal\fi
\begin{document}

\title{A Universal Anti-Spoofing Approach for Contactless Fingerprint Biometric Systems}

\author{Banafsheh Adami, Sara Tehranipoor, Nasser Nasrabadi, and Nima Karimian\\
West Virginia University\\
{\tt\small {ba00011,ft007,nmn0001,nk0033@mix.wvu.edu}}
}

\maketitle
\thispagestyle{empty}

\begin{abstract}
With the increasing integration of smartphones into our daily lives, fingerphotos are becoming a potential contactless authentication method. While it offers convenience, it is also more vulnerable to spoofing using various presentation attack instruments (PAI). The contactless fingerprint is an emerging biometric authentication but has not yet been heavily investigated for anti-spoofing. While existing anti-spoofing approaches demonstrated fair results, they have encountered challenges in terms of universality and scalability to detect any unseen/unknown spoofed samples. To address this issue, we propose a universal presentation attack detection method for contactless fingerprints, despite having limited knowledge of presentation attack samples. We generated synthetic contactless fingerprints using StyleGAN from live finger photos and integrating them to train a semi-supervised ResNet-18 model. A novel joint loss function, combining the Arcface and Center loss, is introduced with a regularization to balance between the two loss functions and minimize the variations within the live  samples while enhancing the inter-class variations between the deepfake and live samples. We also conducted a comprehensive comparison of different regularizations' impact on the joint loss function for presentation attack detection (PAD) and explored the performance of a modified ResNet-18 architecture with different activation functions (i.e., leaky ReLU and RelU) in conjunction with Arcface and center loss. Finally, we evaluate the performance of the model using unseen types of spoof attacks and live data. Our proposed method achieves a Bona Fide Classification Error Rate (BPCER) of 0.12\%, an Attack Presentation Classification Error Rate (APCER) of 0.63\%, and an Average Classification Error Rate (ACER) of 0.37\%. 
\end{abstract}
\vspace{-6pt}

\section{Introduction}
Biometric systems have been used in wide range of applications such as law enforcement and forensics, individual identification, healthcare, and access control for smart phone and tablet which increase convenience in our daily life. Some of the traditional contact-based biometric systems such as fingerprints and palm prints required the physical touch of the user to a sensor, which then increases user concern about hygiene and public shared devices. Moreover, aside from hygiene-related issues, the elasticity of human skin can lead to shape and detail changes in captured touch base biometric when direct contact is made with scanner~\cite{grosz2021c2cl}. Due to this concern and the recent advancement in sensors and cameras, contactless biometrics have gained major popularity for commercial use. Contactless biometric systems also offer high speed authentication/identification since it does not require physical contact. For instance, smartphones can be used to capture finger photos to be used for biometric authentication. However, obtaining high quality image to extract minutia of fingerprints is challenging compared to contact based sensors due to low contrast of ridge and valley patterns~\cite{tan2020towards,lin2018matching}. Furthermore, although contactless biometrics provide a number of advantages as mentioned earlier, they are also more susceptible to deepfake and spoofing attack~\cite{kolberg2023colfispoof}. For instance, facial recognition systems are vulnerable to deepfake such as masking, or contactless fingerprint are susceptible to photopaper or synthetic. 

The presence of various types of attacks, as well as the emergence of unpredictable attack techniques, highlights the need for generalization in presentation attack detection (PAD) systems to effectively detect unseen types of attacks \cite{nikisins2018effectiveness}. Although face biometric anti-spoofing has been extensively studied in the literature, there is currently a lack of research regarding the study of presentation attack detection on contactless fingerprint biometric systems. Furthermore, the current research on PAD heavily relies on supervised learning techniques, where both genuine and spoofed samples are used during the model training -- exhibits poor performance against unseen attacks. To address that limitation, we present a semi-supervised learning model that utilizes a residual network (ResNet18) with training of only live and synthetic spoofed samples. The main contributions of this paper are as follows:

\begin{figure*}
    \centering
    \includegraphics[width=0.95\linewidth]{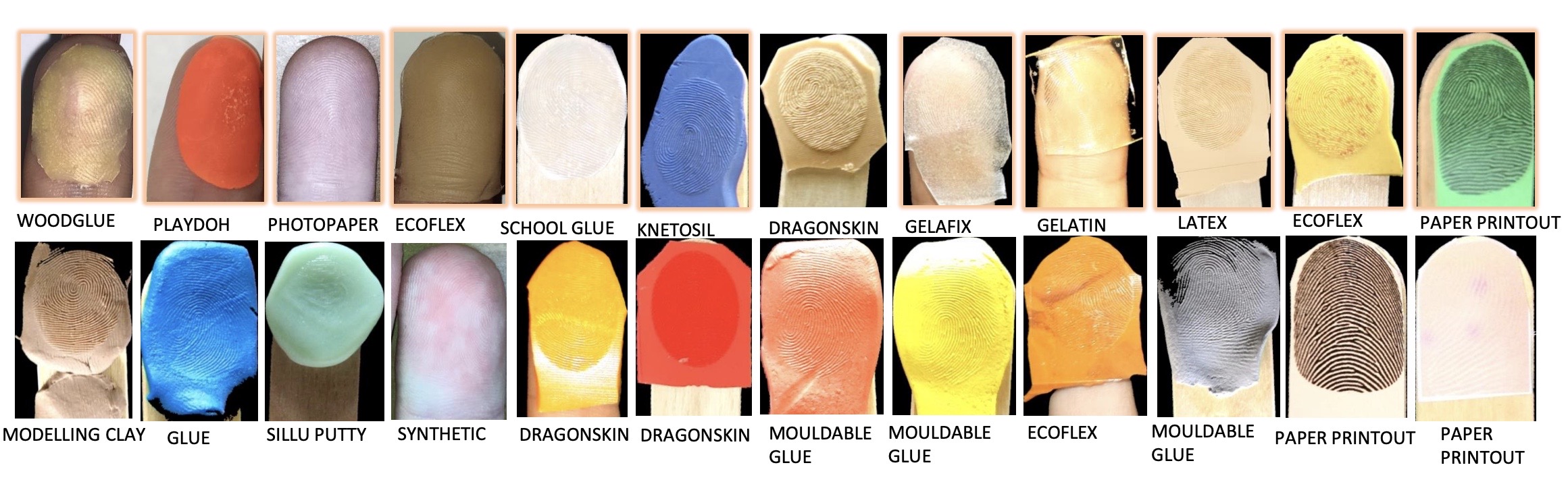}    
    \caption{Different spoofed samples from CLARKSON and COLFISPOOD datasets were employed in this paper.}
    \label{fig:PAI}
\end{figure*}

\begin{itemize}
    \item We proposed a universal presentation attack detection mechanism for contactless fingerprints based on limited knowledge of presentation attack samples. To that end, we generated synthetic contactless fingerptints from live samples using StyleGAN to train a semi-supervised RestNet18. The model is trained on genuine data along with only synthetic spoof attacks. \vspace{-6pt}
    \item We introduced a novel joint loss function by combining the Arcface and Center Loss functions along with regularization to balance between two loss functions. By employing the joint loss function, we aim to minimize the variations within the live samples, while simultaneously enlarging the inter-class variations between deepfake and live samples.\vspace{-6pt}
    \item We conducted a comprehensive comparison of various regularizations' impact on the joint loss function for PAD. Additionally, we evaluated and demonstrated the results of using a modified ResNet-18 architecture with different activation functions, such as leaky ReLU and ReLU, in combination with Arcface and Center loss. Our findings shed light on the effectiveness of these combinations and their performance in the context of anti-spoofing.\vspace{-6pt}
    \item To evaluate our universal PAD, we have stressed out the proposed model with unseen spoofed samples. The model is tested under two public PAI data adopted from COLFISPOOF and CLARKSON. The BPCER and APCER are used as the standard metrics to demonstrate the effectiveness of our proposed technique. The results show that we were able to achieve 0.12\% of BPCER, 0.63\% APCER on various types of spoofed samples. \vspace{-6pt}
    \item We also conducted a comprehensive comparison with state-of-the-art techniques in terms of various scenarios such as evaluation metrics, number of subjects, and detection of unseen attacks under two public datasets. Our proposed work achieved a remarkable 99\% improvement in BPCER at an APCER of 10\% compared to the results in \cite{li2023deep}. Furthermore, when compared to the findings in \cite{purnapatra2023presentation}, our model exhibited substantial improvements of 69.58\% for Photopaper, 1.55\% for Playdoh, and 0.94\% for the synthetic sample
\end{itemize}
\vspace{-2pt}

The paper is structured as follows:
In section \ref{sec:review} we provide an overview of the previous research on contactless fingerprint anti-spoofing detection. In section \ref{sec:proposed}, we present our deep learning architecture used in our experiments. We will demonstrate the database, and experimental set up such as database, evaluation metrics in Section \ref{sec:setup}.  We present and analyze the results obtained from our experiments, comparing them with several approaches found in the literature in Section \ref{sec:setup}. Finally, we conclude our paper in Section \ref{sec:conc}.

\vspace{-6pt}

\section{Related Work}
\label{sec:review}

\begin{table*}[htp]
\begin{center}
\resizebox{\textwidth}{!}{%
\begin{tabular}{|l|l|l|l|l|l|l|l|l|}

\hline
Author & Year & Method & Database & device & Spoof type & Metrics & Results \\ \hline
Tanej et al.~\cite{taneja2016fingerphoto} & 2016 & Hand crafted & \begin{tabular}[c]{@{}l@{}}IIITD: \\ class: 128 \\ images: 5100 \end{tabular} & iphone 5& \begin{tabular}[c]{@{}l@{}}Print Attack\\ Photo Attack\end{tabular}  & \begin{tabular}[c]{@{}l@{}}EER\\ TAR\\ FAR\end{tabular} & EER = 3.71\% \\ \hline
Wasnik et al.~\cite{wasnik2018presentation} & 2018 &\begin{tabular}[c]{@{}l@{}} Hand crafted LBP,\\ BSIF, HOG, SVM \end{tabular}& \begin{tabular}[c]{@{}l@{}} subjects: 50 \\images: 250 \\  videos: 150\end{tabular} & \begin{tabular}[c]{@{}l@{}}IOS, iphone, \\ ipad\end{tabular} & \begin{tabular}[c]{@{}l@{}}print artefact\\ electronic replay\\ elctronic dispaky\end{tabular} & \begin{tabular}[c]{@{}l@{}}APCER\\ BPCER\end{tabular} & \begin{tabular}[c]{@{}l@{}}BPCER = 1.8, 0, 0.66,\\ APCER = 10\end{tabular} \\ \hline
Fujito et al.~\cite{fujio2018face}& 2018 & AlexNet & \begin{tabular}[c]{@{}l@{}}Live: 4096\\ spoofe sample: 8192\end{tabular} & \begin{tabular}[c]{@{}l@{}}ios, android, \\ windows\end{tabular} &  \begin{tabular}[c]{@{}l@{}}Print Attack \\ Photo Attack\end{tabular} & HTER & HTER = 0.04\% \\ \hline
Marasco et al.~\cite{marasco2021fingerphoto}, \cite{marasco2021deep} & 2022 & \begin{tabular}[c]{@{}l@{}}AlexNet DenseNet201,\\  ResNet18,DenseNet121,\\ ResNet34, MobileNEt-V2\end{tabular} & IIITD & \begin{tabular}[c]{@{}l@{}}Apple iphone 5, \\ Flash off, \\ 8MP resolution\end{tabular} & Print AttackPhoto Attack &  \begin{tabular}[c]{@{}l@{}}DEER \end{tabular} & \begin{tabular}[c]{@{}l@{}}D-EER\_AlexNet = 2.14\\ D-EER\_ResNet = 0.97\%\end{tabular} \\ \hline
Kolberg et al.~\cite{kolberg2023colfispoof}& 2023 & Not Reported & \begin{tabular}[c]{@{}l@{}}COLFISPOOF:\\ 7200 spoof samples \\ 72 different PAI\end{tabular} & Not Reported & \begin{tabular}[c]{@{}l@{}}Knetosil, Mould glue, \\ latex, silly putty,\\ paper printout, s\\ chool glue, \\ dragonskin, \\ ecoflex, gelatin, \\ glue,  modelling clau, \\ playdoh\end{tabular}  & Not Reported & Not Reported \\ \hline
Purnapatra et al.~\cite{purnapatra2023presentation} & 2023 & \begin{tabular}[c]{@{}l@{}}DenseNet 121,\\ NASNet\end{tabular} & \begin{tabular}[c]{@{}l@{}}35 subjects with 12 devices\\ attack sample: 7548 \\ synthetic: 10000\end{tabular} & ios, android & \begin{tabular}[c]{@{}l@{}}ecoflex, playdoh, wood glue, \\ synthetic, fingerphoto, latex\end{tabular}  & \begin{tabular}[c]{@{}l@{}}APCER\\ BPCER\end{tabular} &\begin{tabular}[c]{@{}l@{}} APCER = 0.14\% \\ BPCER = 0.18\% \end{tabular}\\ \hline

Hailin Li et al.~\cite{li2023deep} & 2023 & \begin{tabular}[c]{@{}l@{}}AlexNet, DenseNet201,\\ MobileNet-V2, NASNet,\\ ResNet50, GoogleNet,\\ EfficientNet-B0 and\\ Vision Transformers\end{tabular} & \begin{tabular}[c]{@{}l@{}}5886 bonafide and 4247 \\ attack samples with 4 PAIs\end{tabular} & ios, android & \begin{tabular}[c]{@{}l@{}}ecoflex, playdoh, wood glue, \\  fingerphoto, latex\end{tabular}  & \begin{tabular}[c]{@{}l@{}}APCER\\ BPCER\end{tabular} & EER = 8.26\%\\
\hline
\end{tabular}%
}
\caption{Summary of state-of-the-art approaches for contactless fingerprint anti-spoofing. HOG-- histogram of oriented gradients (HOG), SVM-- support vector machine, LBP--local binary
patterns, BSIF--binarized statistical image features, EER -- equal error rate, TAR -- true acceptance rate, FAR -- false acceptance rate BPCER--bonafide presentation classification error rate, HTER -- half total error rate, APCER-- attack presentation classification error rate.}\vspace{-6pt}
\label{tab:previous works}
\end{center}
\end{table*}

The next-generation techniques were deployed based on deep learning approaches in which the model was trained on both live and spoofed samples. Fujio et al.~\cite{fujio2018face} were pioneers in exploring the use of deep neural networks, specifically "Alexnet," for contactless fingerprint anti-spoofing. They trained the model using a combination of bonafide samples and photo-printed attack samples. Their dataset included 9,192 spoofed samples and 4,096 genuine samples, and they achieved an impressive half-error rate of 0.04\%. Marasco et al.~\cite{marasco2021fingerphoto} utilized ResNet and AlexNet architecture based on the IIITD Spoofed Finger photo Database. The database contains 2,048 print attacks, 6,144 photo attacks, and 4,096 live samples. They achieved a D-EER of 2.14\% for AlexNet and .07\% for ResNet, respectively. In subsequent research by Marasco et al.~\cite{marasco2021deep}, there was a slight improvement in the results compared to the baseline approach. However, it should be noted that the ResNet architecture used in their model was trained on both live and spoofed images, which may not be practical or representative of real-world scenarios. Recently, Purnapatra et al.\cite{purnapatra2023presentation} proposed DenseNet-121 and NasNetMobile models with new public database which includes 35 subjects with 65,972 images which includes 29,204 live samples. Their model achieved 88.3\% APCER and 0.48\% BPCER. Furthermore, Hailin Li et al.~\cite{li2023deep} demonstrated PAD using various models such as AlexNet, DenseNet-201, MobileNet-V2, NASNet, RestNet50, and Vision transformer. They have integrated 5,886 live samples and 4,247 spoofed samples and obtained an EER of 8.6\% on RestNest50. Nevertheless, while recent models demonstrated fair performance on spoofed images during training, their generalizability is limited, leading to poor performance when tested on unseen spoofed images. \vspace{-17pt}

\begin{figure*}
    \centering
    \includegraphics[width=0.95\linewidth]{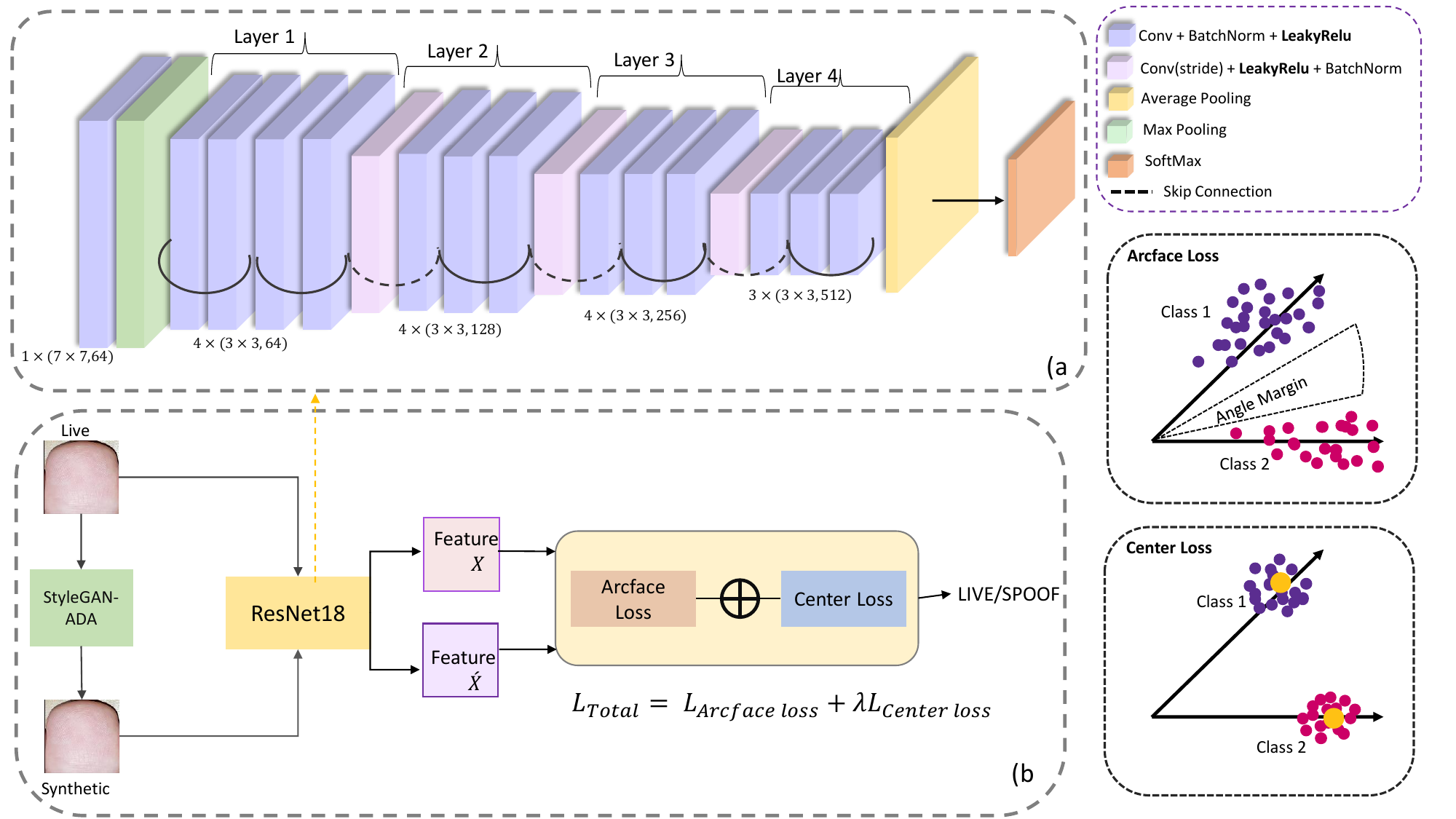}    
    \caption{Resnet-18 with Leaky-Relu activation function and using a joint loss function a combination of the Arcface and center loss functions. The model is trained on live samples and synthetic spoofed fingerphotos and tested under various spoofed images such as photo paper, Ecoflex, playdoh, woodglue, and etc.}\vspace{-6pt}
    \label{fig:model}
\end{figure*}
\section{Proposed Method}
\vspace{-6pt}
\label{sec:proposed}
As illustrated in Figure \ref{fig:model} we applied supervised learning on two fingerprint classes, live and synthetic. In order to increase the resolution and quality of dataset, we applied enhanced super-resolution generative adversarial networks (ESRGAN) on dataset. Additionally, we created synthetic attacks by applying StyleGAN2 with adaptive discriminator augmentation (ADA) \cite{karras2020training} on the live dataset. Furthermore, we employed a pretrained ResNet18 architecture, making some modifications (details of architecture is described in Section \ref{sec:res}). To improve presentation attack detection success rate, we introduced a novel loss function which is a combination of the Arcface loss and Center loss. Our model successfully classified live data samples from unseen spoof type of attacks. In the rest of this section, we will provide a comprehensive explanation of our method and new joint loss function. This method is specifically designed to address contactless fingerprint anti-spoofing. 

\subsection{Joint Loss Function}
In this project, we propose a novel approach for contactless fingerprint anti-spoofing using a joint loss function that combines the Arcface Loss \cite{deng2019arcface} and Center Loss \cite{wen2016discriminative}. Since our approach focouses on universal deepfake detection, utilizing joint loss function will help us to obtain lower error detection rate. By employing the joint loss function, we aim to minimizing the variations within live samples, while simultaneously enlarging the inter-class variation between deep-fake and live samples. Equation \ref{eq:combineloss} demonstrates how our proposed joint function has been calculated by combining Arcfac and center loss.  We used the Arcface loss to map the input data (live and synthetic) to angular space and used center loss to minimizing the variations within each class and pulling together the embeddings of samples belonging to the same class in the feature space: \vspace{-12pt}

\begin{multline*}
    L_{joint} =  -log\frac{e^{scos(\theta_{y_{i}}+m)}}{e^{scos(\theta_{y_{i}}+m)} + \sum_{J=1, J\neq y_{i}}^{N}e^{scos\theta_{y_{j}}}} \\ 
    + \lambda \frac{1}{2}\sum_{i=1}^{m}\left \| x_{i} - c_{y_{i}} \right \|_{2}^{2} \\
\end{multline*}
\vspace{-20pt}
\begin{align}
    L_{joint}  = L_{a} + \lambda L_{c}
    \label{eq:combineloss}
\end{align}

\noindent In above equation, for the first term, N is the number of classes, s is the scaling factor that control the degree between classes, m is additive angular margin, $\theta_{y_{i}}$ is the angle between the feature vector of i-th sample and weight vector of its class, $\theta_{j}$ is the angle between the feature vector pf $i^{th}$ sample and weight vector of $j^{th}$ class. In the second term of above equation m is total number of samples in training in each batch, $x_{i}$ is the feature embedding, $y_{i}$ is the class label of $i^{th}$ sample, and $c_{y_{i}}$ is the center of class. Finally, the $L_{a}$ is Arcface loss, $L_{c}$ is the Center loss, and $\lambda$ is regularization parameter which used to balance the two Arcface and center loss functions. We achieve the best value for lambda during network training.

\subsection{ResNet}
\label{sec:res}
We used ResNet-18 as a deep convolutional Neural Network architecture by applying some modifications to the architecture. We replaced the RELU activation function with Leaky-Relu. The ReLU activation function can encounter a problem called "Dead Neuron" \cite{xu2020reluplex}. Where neurons become inactive for negative inputs, resulting in unchanging weights during training. To tackle this problem, we replace the ReLU activation function with the Leaky-Relu activation function. Using Leaky-Relu addresses this issue by allowing a small, non-zero output for negative inputs, ensuring that neurons don't die out completely and giving an opportunity for weights to be updated during training \cite{maas2013rectifier}. In simpler terms, Leaky-ReLU prevents the dying neuron problem of ReLU by permitting some activity for negative inputs. 

The ResNet-18 contains 5 layers and each layer contains convolutional layers, activation functions, and batch normalization. The input image is $1024\times1024$ with 3 channels (RGB) and fed to the ResNet-18 network. After the first convolutional the number of channels increases to 64, and in the last layer it would have 512 channels. In layers 2, 3, and 4 we have residual blocks in each of them. Each residual block contains two convolutional layers. After the last layer, we have a fully connected layer which is the classification layer, is responsible for converting the learned high-dimensional features from the previous layers into a format suitable for making class predictions. The last block is softmax activation function and it is applied to the output of the last fully connected layer. It takes the single scores for each live and spoof class and converts them into a probability distribution. 
According to Figure \ref{fig:model}.b, we use live and generated synthetic samples from the live dataset as an input to the modified ResNet-18 (See Figure \ref{fig:model}-a). We then apply the Arcface and center loss  on each feature vector and combine them to achieve a better classification.

\begin{table}[htbp]
\centering
\begin{tabular}{cc}
\hline
\multicolumn{2}{c}{SPOOF} \\
\hline
API & NUMBER of IMAGES \\
\hline
ECOFLEX & 1248 \\
PHOTOPAPER & 1104 \\
PLAYDOH & 1700 \\
WOODGLUE & 272 \\
\hline
\multicolumn{2}{c}{LIVE (26 subjects)} \\
\hline
LIVE & 5886 \\
\hline
\hline
\end{tabular}
\caption{Statistics of CLARKSON Dataset.}
\label{tab:clarksondataset}
\end{table}

\begin{table}[htbp]
\centering
\begin{tabular}{c c}
\hline
\textbf{Spoof} & \textbf{Number of Images} \\
\hline
dragonskin & 1700 \\

ecoflex & 300 \\

gelafix & 100 \\

gelatin & 100 \\

glue & 200  \\

knetosil & 200 \\

latex & 100 \\

modelling-clay & 100 \\

mouldable-glue & 900 \\

paper-printout & 1200 \\

playdoh & 1700 \\

silly-putty & 600 \\
\hline
\hline
\end{tabular}
\caption{Statistics of the COLFISPOOF dataset.}
\label{tab:COLFISPOOFDATASET}
\end{table}

\section{Experimental Setup}
\label{sec:setup}
\subsection{Database}
In our research, we made use of two publicly available databases: CLARKSON and COLFISPOOF. The CLARKSON database, introduced by Purnapatra et al. \cite{purnapatra2023presentation}, consists of a variety of images. It includes 7,500 images of four-finger attacks, over 14,000 manually segmented images of single-fingertip attacks, and 10,000 synthetic fingertip images generated using deepfake techniques. These images were obtained from six different Presentation Attack Instruments (PAI) that cover three levels of difficulty. Furthermore, the CLARKSON database contains a total of 31,702 images of 26 subjects captured from live finger photos. Among these images, 2,150 were collected from the four-finger scenario, and 7,768 were collected from single fingertip scenarios. To assess the effectiveness and performance of each device, we evaluated the six different smartphones: iPhone X, iPhone 7, Samsung Galaxy S9, Google Pixel, Samsung Galaxy S6, and S7. For generating spoofed fingertip images, we used different smartphones, such as synthetic, Ecoflex PAI, Playdoh PAI, Wood Glue PAI, Finger Photo PAI, and Latex PAI.

In contrast, the COLFISPOOF database, introduced by Kolberg et al. \cite{kolberg2023colfispoof}, exclusively contains spoof images from various categories, including dragonskin, ecoflex, gelafix, gelatin, glue, knetosil, latex, modelling-clay, moduldable-glue, paper-printout, playdoh, and silly-putty. The statistics of the databases are presented in Table \ref{tab:clarksondataset} and Table \ref{tab:COLFISPOOFDATASET}, showing the details of the CLARKSON and COLFISPOOF datasets, respectively.

\begin{table}[htbp]
\centering
\begin{tabular}{ccc}
\hline
\multicolumn{3}{c}{APCER\%} \\
\hline
API/METHOD & \textbf{ResNet-Leaky-Relu}   & ResNet-Relu \\
\hline
ECOFLEX & \textbf{0}  & 0 \\
PHOTOPAPER & \textbf{9.43}  & 18.60 \\
PLAYDOH & \textbf{0}  & 0 \\
WOODGLUE & \textbf{0}  & 0\\
SYNTHETIC &\textbf{0.15} & 0.25\\
\hline
\multicolumn{3}{c}{BPCER\%} \\
\hline
LIVE & \textbf{0.12}  &0.12 \\
\hline
\hline
\end{tabular}
\caption{Results on CLARKSON dataset (APCER\%, BPCER\%).}
\label{tab:resultClarkson}
\end{table}

\begin{table}[htbp]
\centering
\begin{tabular}{ccc}
    \hline
    \multicolumn{3}{c}{APCER\%} \\
    \hline
    API/METHOD & \textbf{ResNet-LRelu}  & ResNet-Relu \\
    \hline
    DRAGONSKIN & \textbf{0} & 0 \\
    ECOFLEX & \textbf{0} & 0 \\
    GELAFIX &\textbf{0} & 0 \\
    GELATIN & \textbf{0} & 0\\
    GLUE & \textbf{0} & 0\\
    KNETOSIL & \textbf{0} & 0\\
    LATEX & \textbf{0} & 0\\
    MODELLING-CLAY & \textbf{0} & 0\\
    MODULABLE-GLUE & \textbf{0} & 0\\
    PAPER-PRINTOUT & \textbf{0} & 0\\
    PLAYDOH & \textbf{0} & 0\\
    SILLY-PUTTY & \textbf{0} & 0\\
    \hline
    \hline
    \end{tabular}
    \caption{Results on COLFISPOOF dataset (APCER\%). The BPCER for both method is $0.12\%$ for live samples in CLARKSON dataset.}
    \label{tab:resultColfi}
    \end{table}
    
    \begin{table}[htbp]
    \centering
\begin{tabular}{ccc}
\hline
\multicolumn{3}{c}{APCER\%} \\
\hline
Metrics/METHOD & \textbf{ResNet-Leaky-Relu}  & ResNet-Relu \\
\hline
APCER\% &\textbf{0.63} & 1.25 \\
BPCER\% & \textbf{0.12} & 0.12 \\
ACER\% & \textbf{0.37} & 0.68 \\
\hline
\hline
\end{tabular}
\caption{Average of APCER, BPCER, and ACER on both COLFISPOOF and CLARKSON dataset.}
\label{tab:resultClarkson}
\end{table}

It is important to note that the CLARKSON dataset used in our study contains a smaller number of live and spoofed samples compared to the one reported in the aforementioned reference. The original database also includes synthetic spoofed samples that were not investigated in our study.
In order to generate synthetic samples from live fingers, we implement StyleGAN with Adaptive Discriminator Augmentation (ADA) \cite{karras2020training} and generate 5,000 synthetic samples from 26 live subject from CLARKSON dataset. Also, in order to increase the resolutiuon of images, we apply Enhanced Super Resolution Generative Adversarial Networks (ESRGAN)\cite{wang2018esrgan} on both live and synthetic images.

\begin{table*}[htp]
\begin{center}
\resizebox{\textwidth}{!}{%
\begin{tabular}{l|l|l|l|l|l|l}

Method                   & \multicolumn{5}{l|}{(\%)APCER}     & (\%)BPCER \\
\cline{2-6}
                         & ECOFLEX & PHOTOPAPER & PLAYDOH & WOODGLUE & SYNTHETIC&  LIVE   \\
\hline
DenseNet-121~\cite{purnapatra2023presentation}       & 0       & 88.03      & 0.14    & 0     & 0.13      & 0.18  \\
DenseNet-121 (keras) ~\cite{purnapatra2023presentation}     & 0       & 79.01      & 1.55    & 0.94   & 0.79  & 3.64  \\
NasNetMobile~\cite{purnapatra2023presentation}            & 0       & 82.15      & 0.71    & 5.96  & 4.12   & 9.04  \\
DenseNet-121 (grayscale)~\cite{purnapatra2023presentation} & 0.16    & 98.9       & 1.98    & 11    &11.58       & 0.18  \\
ResNet-18/Relu (Combined Loss)       & 0       & 18.6    & 0    & 0     & 0.25     & 0.12  \\
\textbf{Resnet-18/Leaky Relu (Combined Loss)}      & \textbf{0}       & \textbf{9.43}      & \textbf{0}    & \textbf{0}     & \textbf{0.15}      & \textbf{0.12}  \\
ResNet-18/Relu (Arcface Loss)      & 0      & 45.12     & 0    &  0  & 0.32     & 0.37 \\
ResNet-18/Relu (Center Loss)     & 0       & 20.14     & 0   & 0    & 0.26     & 0.26 \\
\hline
\hline
\end{tabular}%
}
\caption{Performance of different deep learning architectures across spoofed and non-spoofed samples, measured in terms of BPCER and APCER.}
\label{tab:finalresults}
\end{center}
\end{table*}

Table~\ref{tab:clarksondataset} and Table~\ref{tab:COLFISPOOFDATASET} shows the detailed statistics of the CLARKSON and COLFISPOOF databases, respectively. Figure~\ref{fig:PAI} shows different spoof types that we used in our study.

\subsection{Implementation Details}
\begin{figure}
    \centering
    \includegraphics[width=1\linewidth]{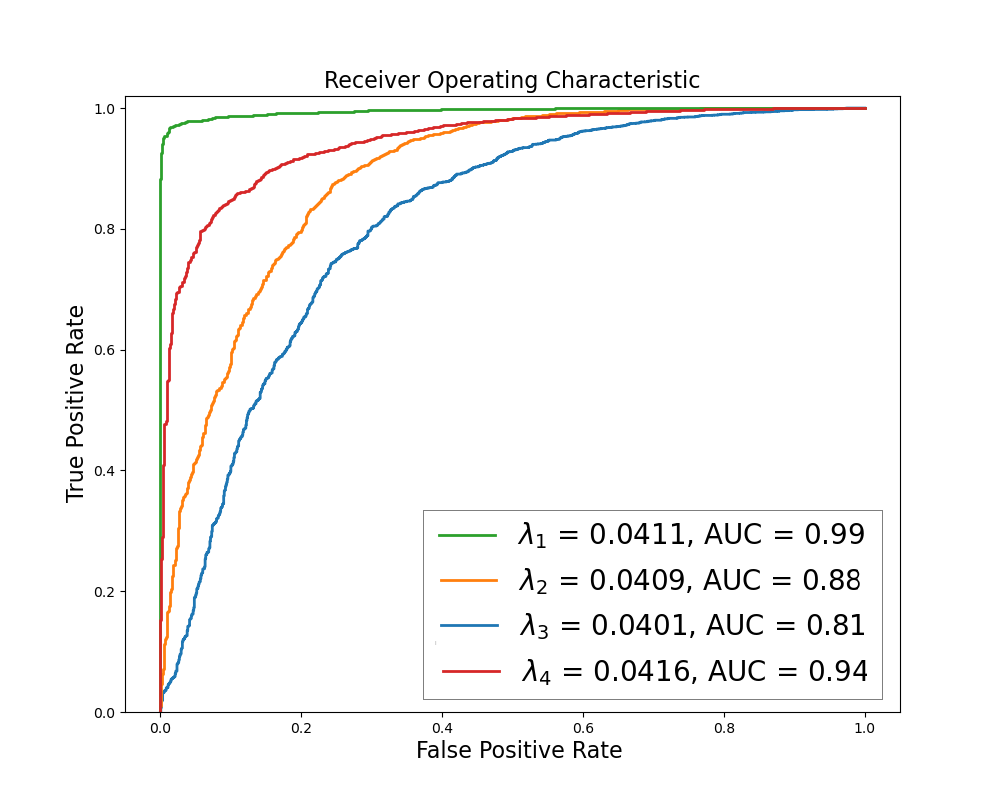}
    \caption{ROC curves for the ResNet architecture trained using live and synthetic samples from StyleGAN, varying the $\lambda$ regularization to find the optimal value.}
    \label{fig:lamda}
\end{figure}

We implemented our combined loss function on ResNet-18 with several modifications, including changing the activation function to Leaky ReLU. The model was trained for 20 epochs using the Adam optimizer with a learning rate of 0.001. Throughout the training process, we assessed the model's performance to determine the optimal value for the $\lambda$ in Equation \ref{eq:combineloss}, which balances the two loss functions. Figure \ref{fig:lamda} illustrates the ROC curve for different $\lambda$ values, and we obtained the best value based on the ROC curve of the validation set. In the Arcface loss function, we set the angle equal to $30^{\circ}$ and added a margin of $0.3$ to the angular.
\subsection{Metrics and Evaluation Protocols}
During the validation phase of our study, we computed the True Positive Rate (TPR) and False Positive Rate (FPR) to determine the probability of correctly classifying live finger images as live and spoof finger images as live images. Additionally, we calculated the Bona Fide Presentation Classification Error Rate (BPCER), Attack Presentation Classification Error Rate (APCER), and Average Classification Error Rate (ACER). ACER is defined as the average of APCER and BPCER, and these metrics help evaluate the performance of our model in distinguishing between live and spoofed finger images.

We trained the RestNet model to determine the optimal $\lambda$ value (regularization parameter) used in the joint loss function (details in Section~\ref{eq:combineloss}, Figure \ref{fig:lamda}). To that end, both live samples and synthetic images generated from StyleGAN were used for binary classification. The goal was to  minimize variations within live samples while enhancing the inter-class variation between deepfake and live samples. Finally, the evaluation of the proposed framework was based on unseen spoofed and live samples. In the testing phase, pretrained model was stressed out on testing datasets, which contains both spoof and live dataset (CLARKSON and CLFISPOOf), and the synthetic samples that we generated using the styleGAN-ADA. Subsequently, we computed APCER (for spoofed samples), and BPCER (for live dataset). By plotting the receiver operating characteristic (ROC) curve and calculating the area under the curve (AUC), we have evaluated the performance of our model. It is important to consider that we evaluate the performance of our model on unseen spoof types of attacks (expect synthetic type) and unseen live subjects. 

\subsection{Results}
\begin{figure}
    \centering
    \includegraphics[width=\linewidth]{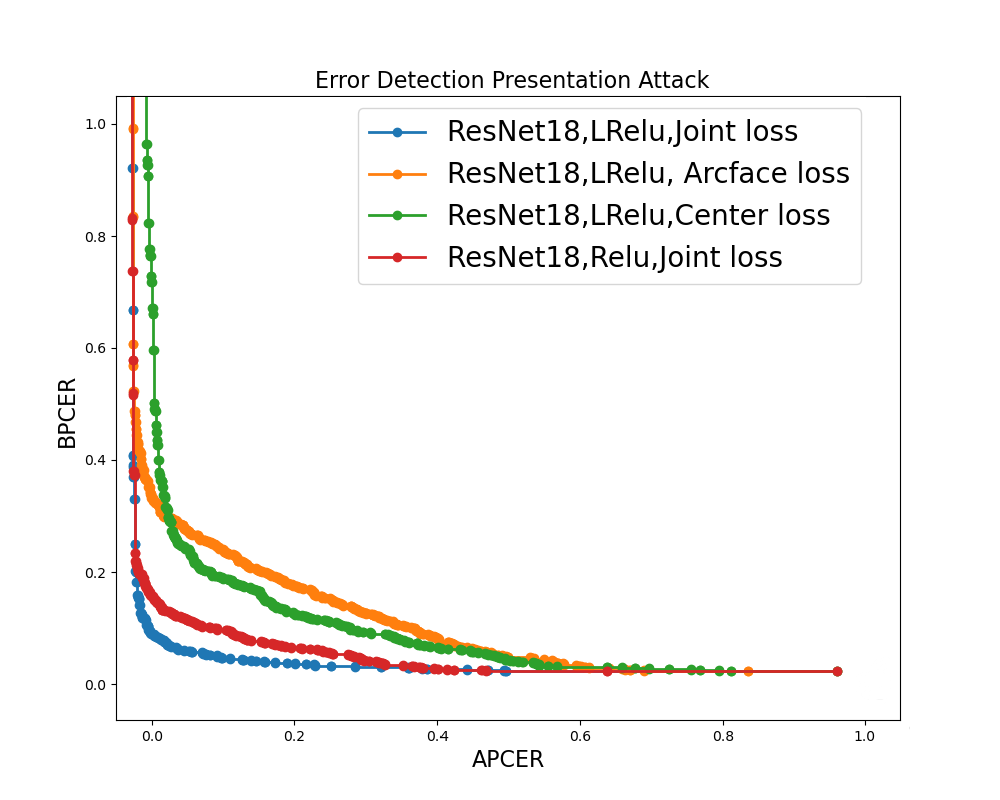}
    \caption{ROC curves illustrate the performance of different loss functions and activation functions using the proposed model.}\vspace{-14pt}
    \label{fig:apcer-bpcer}
\end{figure}

During the training and validation process, we determined the best value for $\lambda$ equal to 0.0411. We tested the trained model on both the CLARKSON and COLFISPOOF datasets, which included live and spoof data. As previously mentioned, we trained the model on a combination of live data (from the CLARKSON dataset) and synthetic data generated using StyleGAN-ADA \cite{karras2020training} from the live dataset. Next, we evaluated the model's performance on unseen spoof attack types from both the CLARKSON and COLFISPOOF datasets, as well as on unseen live subjects. Tables \ref{tab:resultColfi} and \ref{tab:resultClarkson} present the results showcasing the performance of our model on the respective datasets. To provide a better comparison, we trained the ResNet-18 model with both ReLU and LeakyReLU activation functions, utilizing the combined Arcface and Center loss functions. According to Figure \ref{fig:apcer-bpcer}, the ResNet-18 model with LeakyReLU activation function and the combined loss achieved the best performance among the other methods.

\subsection{Discussions}
Based on the results presented in Table \ref{tab:finalresults}, our proposed method, ResNet-18 with Leaky-ReLU and the joint loss function, achieved the best performance compared to other methods in classifying unseen spoof attacks and unseen live datasets. In comparison to DenseNet-121 (keras)\cite{purnapatra2023presentation}, our model's error rate in classifying Photopaper improved by 69.58\%, Playdoh improved by 1.55\%, and for the synthetic samples improved by 0.94\%. Additionally, we achieved an APCER of 0\% for both Ecoflex and Woodglue samples. Furthermore, our model's performance on the live dataset improved by 3.52\%. We also conducted experiments with ResNet-18 using the ReLU activation function and the joint loss function to extract features from images. However, we found that the performance of ResNet with Leaky ReLU was superior. This is because Leaky ReLU addresses the ``dying neuron" problem, which occurs when ReLU-activated neurons in a neural network become stuck and stops learning during training \cite{parisi2020qrelu,maas2013rectifier}. Additionally, we trained ResNet-18 with Leaky ReLU using center loss and Arcface loss independently. According to Table \ref{tab:finalresults}, the modified version of ResNet-18 with Arcface loss improved by 42.91\% and 0.14\% compared to DenseNet-121 in detecting photopaper and playdoh spoof attacks, respectively. However, the results improved even further when applying the Center loss function (refer to Table \ref{tab:finalresults}). ResNet-18 with the center loss function exhibited a 67.89\% improvement in detecting photopaper compared to DenseNet-121 and a 0.14\% improvement in detecting playdoh.

To achieve the best performance, we decided to combine these two loss functions, and as shown in Table \ref{tab:finalresults}, our proposed method achieved the best performance in detecting photopaper, which was the most challenging among the other types of spoof attacks. In addition, in our study, we compared our proposed work with the research conducted by Hailin Li et al.~\cite{li2023deep}. They considered four scenarios in their study, and their ResNet50 architecture was trained as follows: Case-1: Training with photopaper, playdoh, and woodglue, and testing with ecoflex. Case-2: Training with ecoflex, playdoh, and woodglue, and testing with photopaper. Case-3: Training with ecoflex, photopaper, and woodglue, and testing with playdoh. Case-4: Training with ecoflex, photopaper, and playdoh, and testing with woodglue. In our proposed approach, our model was exclusively exposed to synthetic and live samples during training and subsequently tested on all types of unseen spoofed attacks. This strategy allowed us to evaluate the robustness and generalizability of our model against various unseen spoofing scenarios. Based on the findings, in case two where the model is tested on photo-printed attacks, our model demonstrated a remarkable 99\% improvement in BPCER at an APCER of 10\%. This significant improvement highlights the effectiveness of our proposed model in detecting photo-printed attacks.

\section{Conclusion}
The rising popularity of contactless fingerprint biometric systems has led to their potential replacement of conventional touch-based fingerprint recognition systems. However, these systems have some drawbacks, particularly their vulnerability to presentation attacks involving photo-printed or paper printout. Current research in PAD predominantly relies on supervised learning techniques, utilizing both genuine and spoofed samples during training. Nevertheless, these methods often exhibit poor performance against unseen attacks, limiting their scalability. Our deep learning approach is trained solely on genuine images with synthetic data using StyleGAN. During the testing phase, we evaluate our model for all other unseen spoofed samples. We induce a new loss function that combines the Arcface loss to minimize the intra-class variation and the Center Loss to maximize the intra-class variation. By finding the optimal value for a parameter called lambda ($\lambda$), we strike a balance between the two loss functions. Importantly, The proposed scheme demonstrates promising results, with an average BPCER of 0.12\% and an APCER of 0.63\% for presentation attacks on various types of spoofed samples.

\label{sec:conc}
\section{Acknowledgements}
This work was supported in part by the National Science Foundation (NSF) under Grant 2104520.

\end{document}